\documentclass[12pt,times]{elsarticle}
\usepackage{graphicx}
\usepackage{amssymb, amsmath}
\usepackage{url}
\journal{AI Magazine}

\begin{document}

\begin{frontmatter}

\title{Lagrangian based A* algorithm for automated reasoning}
\author{Renju Rajan}
\ead{renjurajan1987@gmail.com}
\address{Department of Basic Science, Muthoot Institute of Technology and Science, Ernakulam, Kerala 682308,
India}

\begin{abstract}
In this paper, a modification of A* algorithm is considered for the shortest path problem. A weightage is introduced in the heuristic part of the A* algorithm to improve its efficiency. An application of the algorithm is considered for UAV path planning wherein velocity is taken as the weigtage to the heuristic. At the outset, calculus of variations based Lagrange's equation was used to identify velocity as the decisive factor for the dynamical system. This approach would be useful for other problems as well to improve the efficiency of algorithms in those areas.
\end{abstract}


\end{frontmatter}

\section{Introduction}
Artificial Intelligence (AI) deals with a set of algorithms that realize automation which do not require human intervention in decision making~\cite{Mue18}. One of the most visible manifestation of artificial intelligence in daily life is in AI cameras in smartphones. In these AI cameras, camera settings are automated based on the scene detected~\cite{Kle14}.  AI camera distinguishes between various scenes such as street, plant, food, text, etc., and captures the image with optimized camera settings which in turn are mapped for each scene. This is refereed to as automated reasoning in artificial intelligence~\cite{Har09}. The level of accuracy in the scene detection varies for different smartphones based on the algorithm that is being employed~\cite{Nix20}. Nevertheless, these AI cameras are capable of providing good quality images based on the preset camera settings that are mapped for each scene.

Decision making in an AI system needs to be spontaneous. This require immediate response to be taken from the system in short notice. This is usually done by resorting to approximate solutions for a given problem. Algorithms written for AI systems are usually approximate solutions. When compared to resource demanding exact solutions, approximate solutions are more suitable for AI systems. For example, Dijkstra's algorithm gives the exact solution for the shortest path problem, whereas A* (A-star) algorithm gives the approximate solution for this problem~\cite{Sab21}. The algorithm which gives approximate solution for a given problem is called heuristic algorithm. Each heuristic algorithm makes use of a heuristic function or rule of thumb to find the approximate solution. In A* algorithm, heuristic function is the straight line distance between the starting and ending points. This is taken as the rule of thumb in arriving at the shortest distance between two points. Heuristic functions are decisive in determining the efficiency of an algorithm. 
\section{Calculus of variations}
Calculus of variations is a mathematical tool which can be used to find the shortest distance between two points on a surface. Any problem in calculus of variations is represented by a definite integral which need to be minimized. 
\subsection{Euler equation}
\begin{figure}[t]
\centering
\includegraphics[width=8cm,height=8cm,keepaspectratio]{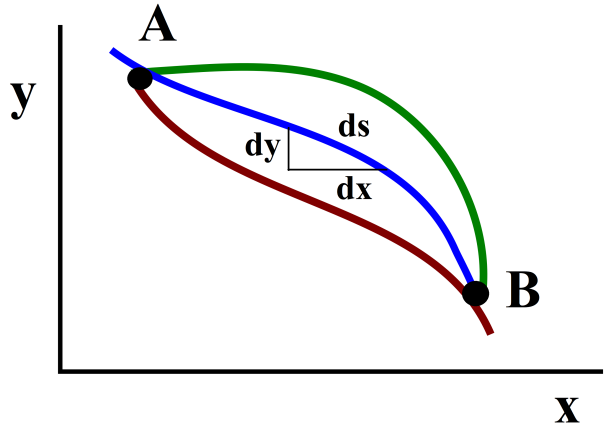}
\caption{Euler equation can be used to find the shortest distance between two points A and B.}\label{fig1}
\end{figure}
According to calculus of variations, it is always possible to write the shortest distance between two points in space as a definite integral, given by,
\begin{equation}\label{Eq.1}
I=\int_{A}^{B} ds.
\end{equation}
As shown in Figure \ref{fig1}, among the various paths, the path which is the shortest gives the minimum value for the definite integral ~\cite{Boa06,Ham14,Van19}. The shortest distance between two points is refereed to as geodesic of the surface.  If the two points fall on a plane, geodesic is a straight line. For points in a plane, it is possible to rewrite Equation \ref{Eq.1} as
\begin{align}\label{Eq.2}
I&=\int_{A}^{B}\sqrt{dx^{2}+dy^{2}}\nonumber\\
&=\int_{A}^{B}f(x,y,y')~dx.
\end{align}
For the definite integral in Equation \ref{Eq.2} to be minimum, a condition need to be satisfied, and it is known as the Euler equation. Euler equation corresponding to Equation \ref{Eq.2} is
\begin{equation}\label{Eq.3}
\frac{d}{dx}\frac{\partial
f}{\partial
y'}-\frac{\partial
f}{\partial
y}=0
\end{equation}
Upon solving Equation \ref{Eq.3}, equation for straight line, $y = mx + c$, is obtained. Straight line is indeed the shortest distance between any two points in a plane. Similarly, for any type of surface, the corresponding geodesic of the surface can be obtained. 
\subsection{Lagrange's equation}
A similar theorem known as the principle of least action holds for mechanical systems wherein the path traversed by the particle from time $t_1$ to $t_2$ is such that the definite integral is a minimum. Here, the definite integral which need to minimized is called action. It is given by
\begin{equation}\label{Eq.4}
I=\int_{t_1}^{t_2}L~dt,
\end{equation}
where $L=T-V$ is the Lagrangian, denoted by difference in kinetic energy and potential energy of the particle. For the definite integral in Equation \ref{Eq.4} to be minimum, a set of conditions known as the Lagrange's equations need to be satisfied. For a particle on the surface of earth, with Cartesian coordinate system, Lagrange's equations are 
\begin{align}\label{Eq.5}
\frac{d}{dt}\frac{\partial
L}{\partial
\dot{x}}-\frac{\partial
L}{\partial
x}=0\nonumber\\
\frac{d}{dt}\frac{\partial
L}{\partial
\dot{y}}-\frac{\partial
L}{\partial
y}=0\nonumber\\
\frac{d}{dt}\frac{\partial
L}{\partial
\dot{z}}-\frac{\partial
L}{\partial
z}=0
\end{align}
Upon solving Equation \ref{Eq.5}, the solutions obtained are equations of motion for the particle, and are given by $\dot{x}=constant$, $\dot{y}=constant$, and $\ddot{z}=-g$.
\section{Lagrangian based A* algorithm}
For objects in motion, Lagrange's equation is suitable for estimating the shortest path. A* algorithm with Lagrangian based heuristic function is considered in this section. 
\subsection{A* algorithm}
A* algorithm modifies Dijkstra's shortest path algorithm by introducing a heuristic. In A* algorithm, each step taken from the current position (current node) is assigned a cost. The aim of A* algorithm is to find the shortest path from start node to the goal node with the lowest cost. The final cost through any node towards the goal node is given by 
\begin{align}\label{Eq.6}
f(n)=g(n)+h(n)
\end{align}
where $f(n)$ is the final cost, $g(n)$ is grid cost and $h(n)$ is heuristic cost. A* algorithm selects the shortest path through the nodes such that the total cost is the lowest in reaching the goal node~\cite{Rex17}.
\begin{figure}[t]
\centering
\includegraphics[width=13cm,height=13cm,keepaspectratio]{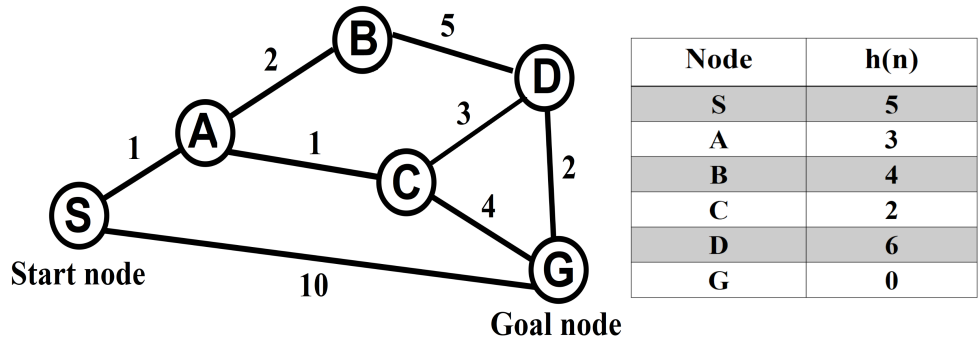}
\caption{A* algorithm makes use of heuristic function to find the shortest path between start node and goal node.}\label{fig2}
\end{figure}

A* algorithm can be illustrated with an example~\cite{Bha19}. Consider nodes as shown in Figure 2. Here, start node is denoted by S and goal node is denoted by G. Shortest path connecting S and G need to be found using A* algorithm. Grid costs are marked in the figure.  Heuristic cost at each node towards the goal node is also given in the table. After calculating the final cost through each path by Eq. \ref{Eq.6}, the shortest path between S and G can be determined. The path with lowest cost would be the shortest. Starting from S, there are two paths to goal node, viz. S$\rightarrow$A and S$\rightarrow$G. One directly reaches goal node G and the other reaches node A. Final costs at A and G nodes are calculated as 4 and 10 respectively. S$\rightarrow$G is kept on hold, and then proceed with S$\rightarrow$A as the cost is lower through that path. Going forward, there are two paths through node A, viz. S$\rightarrow$A$\rightarrow$B and S$\rightarrow$A$\rightarrow$C. Final cost through path S$\rightarrow$A$\rightarrow$B is 7 and through path S$\rightarrow$A$\rightarrow$C is 4. As the path S$\rightarrow$A$\rightarrow$C costs less, path through node C is selected, keeping path S$\rightarrow$A$\rightarrow$B on hold. Moving forward, there are two paths through node C, viz. S$\rightarrow$A$\rightarrow$C$\rightarrow$G and S$\rightarrow$A$\rightarrow$C$\rightarrow$D. Final cost through path S$\rightarrow$A$\rightarrow$C$\rightarrow$G is 6 and through path S$\rightarrow$A$\rightarrow$C$\rightarrow$D is 11. As path S$\rightarrow$A$\rightarrow$C$\rightarrow$G has lower cost, it is selected. Comparing with path S$\rightarrow$G which was put on hold earlier, path S$\rightarrow$A$\rightarrow$C$\rightarrow$G has lower cost again, making it the shortest path between nodes S and G.
\subsection{Lagrangian based heuristic function}
Heuristic function calculates the distance between current node and the goal node. It appears as part of the final cost in A* algorithm (Eq.\ref{Eq.6}). There are many heuristic functions that are commonly used in A* algorithm, viz. Manhattan, Euclidean, and Diagonal, etc. For objects moving with a velocity, it would be appropriate to use a heuristic with velocity as a factor. It is possible to frame a heuristic function based on equations of motion which in turn are obtained by solving Lagrange's equations (Eq.\ref{Eq.5}). As velocity and acceleration are vector quantities, the resultant vector can be used for realizing a heuristic function. 
\section{Application in UAV path planning}
\begin{figure}[t]
\centering
\includegraphics[width=13cm,height=13cm,keepaspectratio]{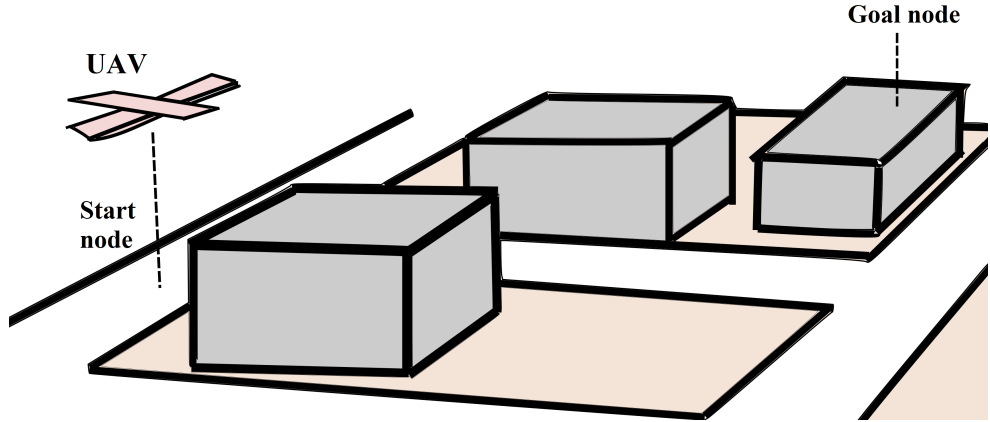}
\caption{UAV needs to circumvent the obstacles to reach the goal node.}\label{fig3}
\end{figure}
One of the applications of A* algorithm is in UAV path planning~\cite{Man21}. Path planning guides the UAV through the shortest path from start node to the goal node by circumventing the obstacles in the path as shown in Figure \ref{fig3}. Since the UAV is moving with a particular velocity, resultant velocity of the UAV in the direction of goal node is an ideal heuristic for calculating the shortest path. In this approach, in contrast with the usual procedure in A* algorithm, velocity will be taken as a weightage to the heuristic cost. Accordingly, the final cost expression takes the form, 
\begin{align}\label{Eq.7}
f(n)=g(n)+vh(n)
\end{align}
where $v$ is the resultant velocity of the UAV in the direction of goal node. This is in line with the understanding that a higher velocity would make the UAV reach the destination faster. As the UAV is moving through a three dimensional space, altitude of the goal node will also need to be taken into account while framing the algorithm. Efficiency of an algorithm is very much determined by the heuristic function that is being employed. Different heuristic functions can be compared for efficiency before selecting the most appropriate one for the algorithm.
\section{Conclusion}
AI systems are expected to make decisions instantaneously. Faster decision making or automated reasoning is achieved with the help of efficient algorithms. There is a continuous effort from the part of researchers to make the algorithm more efficient. A* algorithm discussed is the de facto algorithm for shorted path problem. A modification of A* algorithm is considered wherein velocity is taken as a weightage to the heuristic part of the algorithm. The approach is useful for applications such as UAV path planning. The calculus of variation approach discussed in the initial section can be extended for more types of problems for improving the efficiency of algorithms in those areas. 
\section*{Acknowledgments}
I thank Infinix mobile for kindling my interest in artificial intelligence through AI cameras in their smartphones.

\end{document}